\documentclass{article}

\usepackage{arxiv}

\usepackage[utf8]{inputenc} 
\usepackage[T1]{fontenc}    
\usepackage{hyperref}       
\usepackage{url}            
\usepackage{booktabs}       
\usepackage{amsfonts}       
\usepackage{nicefrac}       
\usepackage{microtype}      

\usepackage{graphicx}
\usepackage{natbib}
\usepackage{doi}

\usepackage{amsthm}
\theoremstyle{definition}

\usepackage{ulem}
\usepackage{standalone}
\usepackage{pifont}
\usepackage{tikz}
\usepackage{subcaption}
\usepackage{enumitem}
\setlist[enumerate]{label*=\arabic*.}
\usetikzlibrary{shapes}
\usepackage{amsmath}

\title{L'explicabilité au service de l'extraction de connaissances : application à des données médicales}

\author{ \href{https://orcid.org/0000-0001-9316-0617}{\includegraphics[scale=0.06]{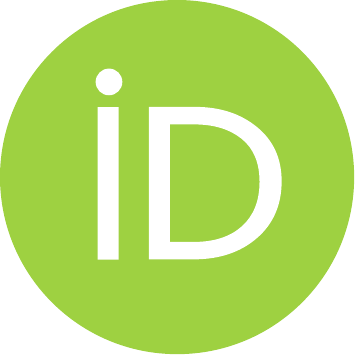}\hspace{1mm}Robin Cugny}
    \\
	IRIT / SolutionData Group\\
	Université Toulouse 2\\
	Toulouse, France \\
	\texttt{robin.cugny@irit.fr} \\
	\And
	\href{https://orcid.org/0000-0003-1185-9630}{\includegraphics[scale=0.06]{orcid.pdf}\hspace{1mm}Emmanuel Doumard} \\
	IRIT / RESTORE\\
	Université Toulouse 3\\
	Toulouse, France \\
	\texttt{emmanuel.doumard@irit.fr} \\
	\And
	\href{https://orcid.org/0000-0003-3618-967X}{\includegraphics[scale=0.06]{orcid.pdf}\hspace{1mm}Elodie Escriva} \\
	IRIT / Kaduceo\\
	Université Toulouse 1\\
	Toulouse, France \\
	\texttt{elodie.escriva@irit.fr} \\
	\And
	\href{https://orcid.org/0000-0002-9624-3212}{\includegraphics[scale=0.06]{orcid.pdf}\hspace{1mm}Haomiao Wang} \\
	RESTORE\\
    Université Toulouse 3\\
	Toulouse, France \\
	\texttt{haomiao.wang@inserm.fr} \\
}

\renewcommand{\shorttitle}{L'explicabilité au service de l'extraction de connaissances}

\hypersetup{
pdftitle={L'explicabilité au service de l'extraction de connaissances},
pdfsubject={cs.LG},
pdfauthor={Robin Cugny, Emmanuel Doumard, Elodie Escriva, Haomiao Wang},
pdfkeywords={Explainable machine learning, Knowledge extraction, Data pipeline, Feature selection, Clustering of explanations, Information system},
}

\begin{document}
\maketitle

\begin{abstract}
The use of machine learning has increased dramatically in the last decade. The lack of transparency is now a limiting factor, which the field of explainability wants to address. Furthermore, one of the challenges of data mining is to present the statistical relationships of a dataset when they can be highly non-linear. One of the strengths of supervised learning is its ability to find complex statistical relationships that explainability allows to represent in an intelligible way.
This paper shows that explanations can be used to extract knowledge from data and shows how feature selection, data subgroup analysis and selection of highly informative instances benefit from explanations. We then present a complete data processing pipeline using these methods on medical data.

\begin{center}
\textbf{Résumé}
\end{center}

L'utilisation de l'apprentissage automatique a connu un bond cette dernière décennie. Le manque de transparence est aujourd'hui un frein, que le domaine de l'explicabilité veut résoudre. Par ailleurs, un des défis de l'exploration de données est de présenter les relations statistiques d'un jeu de données alors que celles-ci peuvent être hautement non-linéaires. Une des forces de l'apprentissage supervisé est sa capacité à trouver des relations statistiques complexes que l'explicabilité permet de représenter de manière intelligible.
Ce papier montre que les explications permettent de faire de l'extraction de connaissance sur des données et comment la sélection de variables, l'analyse de sous-groupes de données et la sélection d'instances avec un fort pouvoir informatif bénéficient des explications. Nous présentons alors un pipeline complet de traitement des données utilisant ces méthodes pour l'exploration de données médicales.
\end{abstract}


\section*{Remerciements}
Nous remercions l'ANRT pour les thèses CIFRE n° [2020/0870] et [2020/0964] ainsi que les compagnie SolutionData Group et Kaduceo, le Programme d’Investissements d’Avenir et l'ANR pour la thèse EUR CARe N°ANR-18-EURE-0003 ainsi que l'allocations doctorales interdisciplinaires N°ALDOCT89533–2021-2026 de l'Université Fédérale Toulouse Midi-Pyrénées et de la Région Occitanie.

\section{Introduction}

Le domaine de l'apprentissage automatique a connu un succès retentissant cette dernière décennie. Toutefois, l'augmentation des performances des modèles s'accompagne d'une augmentation de leur complexité, à la fois en terme de nombres de paramètres mais aussi de quantité de données nécessaires pour les entraîner. Cela entraîne souvent un manque de transparence des opérations effectuées et résultats obtenus, effet appelé "boîte noire". Ce phénomène soulève la question de l'explicabilité des modèles entraînés et se révèle être bloquant dans des domaines à forts enjeux comme la finance, la santé, la justice ou les véhicules autonomes par exemple. L'explicabilité des modèles (aussi appelé XAI, \textit{eXplainable Artificial Intelligence} en anglais) soulève également des problématiques éthiques, légales et sociétales. Ainsi, en Europe, le Règlement Général sur la Protection des Données demande à ce qu'une décision prise par un tel algorithme puisse être expliquée à la personne concernée. 

Dans ce contexte, plusieurs approches d'explicabilité ont vu le jour : les explications globales du modèle, les explications locales des prédictions, les approches basées sur les exemples ou encore les modèles d'apprentissage automatique naturellement interprétables \citep{molnar_2019}. 
Chacune de ces approches met en avant des caractéristiques spécifiques du modèle entraîné, des prédictions effectuées ou du jeu de données utilisé pour l'entraînement.
Les approches proposant des explications locales comme LIME \citep{ribeiro2016should} et SHAP \citep{lundberg_shap_2017} sont celles qui suscitent le plus de propositions scientifiques et sont les plus populaires, autant dans la recherche que l'industrie \citep{nauta2022anecdotal}. Ces approches présentent l'avantage d'être facilement utilisables et compréhensibles par des utilisateurs non-experts de l'apprentissage automatique puisque les explications ont une forme courte, logique et contrastive \citep{molnar_2019}.
Les papiers définissant les méthodes explicatives utilisent néanmoins l'explicabilité comme une finalité, pour justifier de la possibilité d'expliquer le modèle proposé, sans aller plus loin dans l'étude des explications comme objets informatifs. 
Des comparaisons de plusieurs méthodes locales d'explications tendent à définir dans quel contexte chaque méthode est la meilleure \citep{doumard-localXAI-comparaison_2022}. Ces études ont permis une démocratisation de leur utilisation via des exemples d'utilisation, qui essayent d'aller plus loin que les simples explications, notamment au travers de clustering \citep{excoffier_clustering-xai_2022,lee_xai-clustering_2022}. En amont, une autre approche prend en compte les explications pour améliorer la sélection de variables et les étapes de pré-traitement des données \citep{haomiao_explainai2022}. Ces techniques sont cependant utilisées indépendamment les unes des autres, sans enrichir un processus global d'extraction de connaissance.

L'objectif de ce papier est alors de montrer que l'explicabilité peut être bénéfique à plusieurs étapes d'un pipeline d'extraction de connaissances sur un jeu de données. 
Nous présenterons d'abord le pipeline proposé ainsi que ses différents composants.
Nous présenterons et discuterons ensuite les résultats avant de présenter de potentielles perspectives de travail.

\section{Méthodes}

\subsection{Description du pipeline proposé}

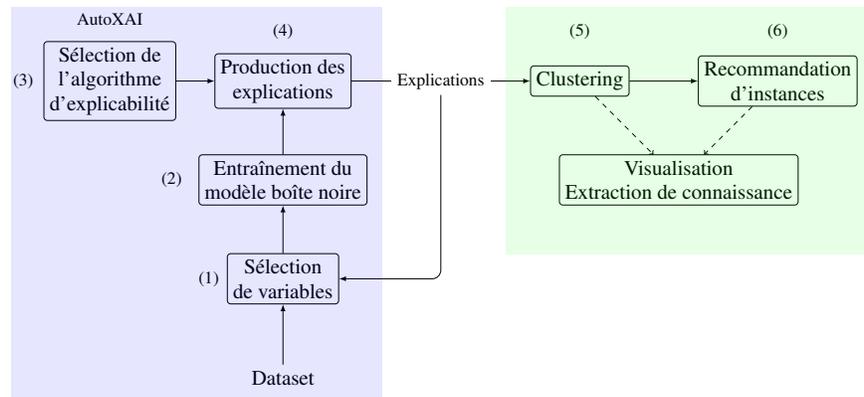
\begin{figure*}[t!]
\centering
\begin{tikzpicture}[every text node part/.style={align=center}]

\node[font=\large] (df) at (0,-6){Dataset};

\node[draw, rounded corners=2pt, font=\large] (sa) at (0,-4){Sélection\\de variables};
\node[] (user0) at (-1.5,-4){(1)};

\node[draw, rounded corners=2pt, font=\large] (bn) at (0,-2){Entraînement du\\modèle boîte noire};
\node[] (user0) at (-2.25,-2){(2)};

\node[draw,rounded corners=2pt,font=\large] (auto) at (-3.5,0){
Sélection de\\l'algorithme\\d'explicabilité};
\node[] (user0) at (-5.25,0){(3)};
\node[] (user0) at (-3.5,1.25){AutoXAI};

\node[draw, rounded corners=2pt, font=\large] (axai) at (0,0){Production des\\explications};
\node[] (user0) at (0,1){(4)};
\node[draw, rounded corners=2pt, font=\large] (clust) at (6,0){Clustering};
\node[] (user0) at (6,1){(5)};

\node[draw, rounded corners=2pt, font=\large] (reco) at (10,0){Recommandation\\d'instances};
\node[] (user0) at (10,1){(6)};

\node[draw, rounded corners=2pt, font=\large] (viz) at (8,-2){Visualisation\\Extraction de connaissance};

\draw[-latex] (df) -- (sa);
\draw[-latex] (sa) -- (bn);
\draw[-latex] (bn) -- (axai);
\draw[-latex] (auto) -- (axai);
\draw[-latex] (axai) -- node [midway, font=\normalsize, fill=white] (exp) {Explications} (clust);

\draw[-latex] (clust) -- (reco);
\draw[->,dashed] (reco) -- (viz);
\draw[->,dashed] (clust) -- (viz);

\draw[->,>=latex,rounded corners=5pt] (exp) |- (sa);

\fill [opacity=0.1,color=blue] (-5.5,-6.5,0)--(2,-6.5,0)--(2,1.5,0)--(-5.5,1.5,0)--cycle ;
\fill [opacity=0.1,color=green] (4.5,-3.5,0)--(12,-3.5,0)--(12,1.5,0)--(4.5,1.5,0)--cycle ;

\end{tikzpicture}
\caption{Pipeline d'extraction de connaissances au travers des explications}
\label{fig:pipeline}
\end{figure*}
Nous commençons dans un premier temps par présenter une vue d'ensemble du pipeline proposé en décrivant étape par étape les opérations effectuées.
Le pipeline de traitement est illustré en Figure \ref{fig:pipeline} et les étapes sont les suivantes :
\begin{enumerate}
    \item Premièrement, les variables pertinentes pour le modèle sont sélectionnées dans le jeu de données.
    \item Le modèle d'apprentissage est entraîné sur le jeu de données constitué du sous-ensemble de variables.
    \item AutoXAI utilise le modèle et le jeu de données pour proposer un algorithme d'explicabilité dont les hyperparamètres sont optimisés.
    \item L'algorithme d'explicabilité va produire des explications pour le modèle boîte noire relativement au jeu de données.
    \item Un algorithme de clustering est appliqué sur ces explications afin de trouver des groupes d'instances similaires de par leurs explications. Chaque cluster est résumé avec un ensemble de règles.
    \item Enfin, on sélectionne les instances les plus informatives sur la distribution des données de chaque cluster pour les présenter à l'utilisateur et faciliter l'étude des données.
\end{enumerate}

La sélection de variables est le processus qui identifie le sous-ensemble de variables le plus pertinent pour construire un modèle d'apprentissage automatique. Cette étape de pré-traitement permet d'améliorer la performance du modèle, d'économiser des ressources liés à l'entraînement du modèle et de faciliter la compréhension des données et de la prédiction \citep{guyon2003introduction}. Dans les applications réelles, comme les variables pertinentes sont généralement inconnues \textit{a priori}, les méthodes sont fréquemment évaluées par des métriques indirectes comme la précision ou le rappel \citep{yu2004fs_indirect}. La sélection de variables utilisée a un impact sur l'explication ce qui donne des indications sur la qualité de cette sélection de variable. Nous considérons que l'explication générée avec toutes les variables porte l'information complète et peut être utilisée comme base de référence. Si une explication générée par un sous-ensemble de variables est alors similaire à la base de référence, ce sous-ensemble est capable de conserver le maximum d'informations et la méthode qui propose ce sous-ensemble est préférable.

Pour la sélection de l'algorithme d'explicabilité, nous faisons appel à l'outil AutoXAI \citep{cugny-autoXAI_2022}. AutoXAI sélectionne les algorithmes d'explicabilité qui produisent des explications correspondant aux besoins de l'utilisateur. Il utilise ensuite des mesures d'évaluation des explications pour optimiser les hyperparamètres des algorithmes d'explicabilité afin de s'assurer que les explications respectent des contraintes de qualité. Ainsi, AutoXAI produit un classement où les algorithmes d'explicabilité sont ordonnés selon leurs scores obtenus par les mesures d'évaluation. Nous pouvons alors choisir un algorithme d'explicabilité produisant des explications adaptées et ayant de bons scores pour les mesures d'évaluations choisies.

Dans le cadre de l'exploration de données, il est commun de chercher à identifier des groupes d'instances parmi les données. Le domaine du clustering cherche à répondre à cette problématique de manière non-supervisée. Les explications offrent un nouvel espace de données pouvant être utilisé pour identifier des sous-groupes d'instances. On minimise ainsi l'impact des variables pouvant s'apparenter à du bruit, et on maximise l'impact des variables importantes pour le modèle.
Cela permet de mettre en valeur des motifs spécifiques à la tâche d'intérêt~\citep{cooper2021supervised}. De plus, le clustering hiérarchique nous permet de raisonner sur des groupes imbriqués, afin d'offrir une plus grande granularité dans l'exploration des populations. 
De plus, nous avons aussi utilisé Skope-rules afin de produire des règles résumant chaque cluster.

La recommandation d'instances permet ensuite de proposer à l'utilisateur un groupe réduit d'instances ayant un fort pouvoir représentatif. À la différence des systèmes classiques de recommandation, cette étape du pipeline utilise exclusivement les explications pour définir les instances les plus importantes à étudier. Cette sélection se base sur les clusters créés précédemment, en sélectionnant l'instance centrale de chaque cluster, le médoïde si disponible, le centroïde sinon. Leur étude peut ainsi permettre de mieux comprendre le jeu de données et potentiellement les instances correctement et incorrectement classées par le modèle.

\subsection{Protocole expérimental}

Nous illustrons notre approche avec \textit{sa-heart}, un jeu de données de classification binaire issu d'une étude rétrospective des hommes d'une région à haut risque de maladies cardiaques de l'Afrique du Sud  \citep{rossouw1983coronary}. Le modèle d'apprentissage utilisé est le \textit{MLPClassifier}, un perceptron multicouche issu de la bibliothèque scikit-learn \citep{scikit-learn}.
L'algorithme d'explicabilité Kernel SHAP est sélectionné via AutoXAI, avec une optimisation des hyperparamètres au regard de la \textit{Robustesse} \citep{alvarez2018robustness} et de l'\textit{Infidélité} \citep{yeh2019fidelity}, deux mesures d'évaluation des explications. 

Pour la sélection de variables, 13 méthodes sont testées pour couvrir les différentes familles de la sélection de variables. En comptant la précision, le $\tau$ de Kendall entre les rangs de variables et le changement de l'influence des variables, peu de méthodes éliminent des variables. Les deux méthodes basées sur la théorie de l'information, \textit{Conditional Mutual Information (cmim)} \citep{fleuret2004cmim} et \textit{Joint Mutual Information (jmi)}\citep{yang1999jmi}, proposent un sous-ensemble commun avec une légère amélioration du taux de bonnes prédictions.

L'algorithme de clustering utilisé sur les explications est hiérarchique et agglomératif, il utilise le critère de Ward~\citep{ward1963application} implémenté dans la bibliothèque Scipy. Pour déterminer le nombre de clusters, nous avons privilégié la méthode L \citep{salvador_determining_2004}, basé sur une automatisation de la méthode du coude pour le clustering hiérarchique.
Afin de faire de la recommandation d'instance, nous identifions l'instance la plus proche du centre de chaque cluster. Ces instances sont aussi appelées médoïdes.

Le code pour reproduire les résultats expérimentaux est disponible à l'adresse suivante : \url{https://github.com/RobinCugny/XAI-knowledge-extraction}

\section{Résultats et discussion}

Dans un premier temps, nous entraînons le modèle d'apprentissage sur le jeu de données complet et son taux de bonnes prédictions est de $0.766$. Nous entraînons le modèle sur toutes les données afin de capturer au mieux toutes les relations statistiques sans nous préoccuper du sur-apprentissage, puisque nous ne cherchons pas à généraliser le modèle.
Vient alors la question suivante : \textit{"Toutes les relations statistiques sont-elles significatives ?"}. En effet, on ne peut exclure la présence potentielle de variables redondantes ou inutiles (bruit) dans le jeu de données. Il convient alors d'effectuer une étape de sélection de variables. Après calculs, il s'avère que la variable \textit{'type-A behavior'} n'apporte que du bruit au modèle d'apprentissage puisque sa précision est de $0.768$ sans celle-ci. Nous décidons donc de l'écarter car sa contribution n'aurait pas été significative voire trompeuse lors de l'extraction de connaissance, puisque le modèle d'apprentissage est perturbé par cette variable.
Nous pouvons à présent utiliser AutoXAI pour choisir l'algorithme d'explicabilité que nous allons utiliser. Un extrait du classement produit par AutoXAI est visible dans le Tableau \ref{table:autoxai_classement}. 

Les algorithmes d'explicabilité sont classés dans l'ordre décroissant d'après leur score agrégé, ce dernier est obtenu par combinaison linéaire des mesures d'évaluation standardisées et les hyperparamètres utilisés sont dans la colonne de droite. 
Les hyperparamètres sont \textit{nsamples}, \textit{l1\_reg} et \textit{summarize} pour Kernel SHAP et \textit{num\_samples} pour LIME.
Dans ce classement, Kernel SHAP est meilleur que LIME puisque la meilleure combinaison d'hyperparamètres pour LIME est seulement $32^{eme}$. Cependant, les 2 meilleures solutions globales n'utilisent que 2 variables, ce qui peut en effet maximiser les mesures d'évaluation mais qui s'avère contre productif pour faire de l'exploration de données. Dans notre cas, nous souhaitons voir l'influence de chacune des variables que nous avons conservées à l'issue de l'étape de sélection de variables. 
La solution que nous choisissons est donc la $3^{eme}$ solution, Kernel SHAP avec les hyperparamètres suivant : 'l1\_reg': "bic", 'nsamples': 581, 'summarize': "KernelExplainer". 

\begin{table}[]
\centering
\small
\caption{Extrait du classement produit par AutoXAI.}
\resizebox{0.9\columnwidth}{!}{
\begin{tabular}{cccccc}
\hline
Rang & \multicolumn{1}{c|}{\begin{tabular}[c]{@{}c@{}}Score\\ agrégé\end{tabular}} & \begin{tabular}[c]{@{}c@{}}Robustesse\\ standardisée\end{tabular} & \multicolumn{1}{c|}{\begin{tabular}[c]{@{}c@{}}Fidélité\\ standardisée\end{tabular}} & \begin{tabular}[c]{@{}c@{}}Algorithme\\ d'explicabilité\end{tabular} & Hyperparamètres \\ \hline
1 & \multicolumn{1}{c|}{1.214} & 0.320 & \multicolumn{1}{c|}{1.054} & SHAP & 1568 ; num\_features(2) ; KernelExplainer \\
2 & \multicolumn{1}{c|}{1.214} & 0.320 & \multicolumn{1}{c|}{1.054} & SHAP & 1731 ; num\_features(2) ; KernelExplainer \\
3 & \multicolumn{1}{c|}{0.879} & 0.007 & \multicolumn{1}{c|}{0.875} & SHAP & 581 ; bic ; KernelExplainer \\ \hline
 &  &  & ... &  &  \\ \hline
32 & \multicolumn{1}{c|}{-0.138} & -1.872 & \multicolumn{1}{c|}{0.798} & LIME & 10 \\
37 & \multicolumn{1}{c|}{-0.520} & -0.603 & \multicolumn{1}{c|}{-0.218} & LIME & 69
\end{tabular}}
\label{table:autoxai_classement}
\end{table}


\begin{table}[]
\caption{Table de règles associées à chaque cluster}
\centering
\resizebox{0.9\columnwidth}{!}{
\centering
\begin{tabular}{cccc}
\hline
\textbf{Cluster} & \textbf{Règles}                                                                                                                                                  & \textbf{\begin{tabular}[c]{@{}c@{}}Taux de bonnes\\ classifications\end{tabular}} & \textbf{Moyenne prédiction} \\ \hline
1                & \begin{tabular}[c]{@{}c@{}}cumulative tobacco \textgreater 9.825\\ obesity \textgreater 19.385\\ age \textgreater 42.0\end{tabular}                            & 73.3\%                                                                            & 66.7\%                      \\
\hline
2                & \begin{tabular}[c]{@{}c@{}}cumulative tobacco $\leq$ 10.85\\ low density lipoprotein cholesterol \textgreater 5.065\\ age \textgreater 37.5\end{tabular} & 67.5\%                                                                            & 52.5\%                      \\
\hline
3                & \begin{tabular}[c]{@{}c@{}}low density lipoprotein cholesterol $\leq$ 4.76\\ obesity $\leq$ 24.33\\ age \textgreater 43.5\end{tabular}             & 71.4\%                                                                            & 40.8\%                      \\
\hline
4                & \begin{tabular}[c]{@{}c@{}}current alcohol consumption $\leq$ 61.815\\ age $\leq$ 40.5 \\ age \textgreater 22.0\end{tabular}                       & 79\%                                                                              & 20.3\%                      \\
\hline
5                & \begin{tabular}[c]{@{}c@{}}obesity $\leq$ 27.185\\ age $\leq$ 23.5\end{tabular}                                                                    & 96.7\%                                                                            & 3.28\%                      \\ \hline
\end{tabular}
}
\label{tableau:regles_clusters}
\end{table}

Avec les explications ainsi produites, nous pouvons appliquer l'algorithme de clustering sur les explications. Les clusters obtenus sont décrits par des règles qui délimitent les contours des clusters à la manière d'un arbre de décision. Ces règles sont visibles dans le Tableau \ref{tableau:regles_clusters}. Pour chaque cluster, on peut voir le taux de bonnes prédictions du modèle sur ce cluster ainsi que le pourcentage de patients malades. Ces informations peuvent renseigner aussi bien un data scientist qu'un expert métier. En effet, les performances du modèle montrent que certaines données sont plus difficiles à modéliser, ce qui indique des modifications à effectuer sur le modèle ou les données et des précautions à prendre pour une application dans un contexte réel. On peut également considérer cet écart de performance comme un biais, ce qui peut soulever des questions éthiques. Ici par exemple, le modèle a de meilleures performances sur les patients de moins de 23 ans par rapport aux autres patients, même si leur nombre est faible. Pour un expert métier, ces clusters peuvent représenter les tendances statistiques majeures de la population étudiée. Avec ces règles, il dispose en plus des relations simplifiées en rapport avec le sujet de l'étude, ici les maladies cardio-vasculaires.

\begin{figure}
\includegraphics [width=0.49\columnwidth]{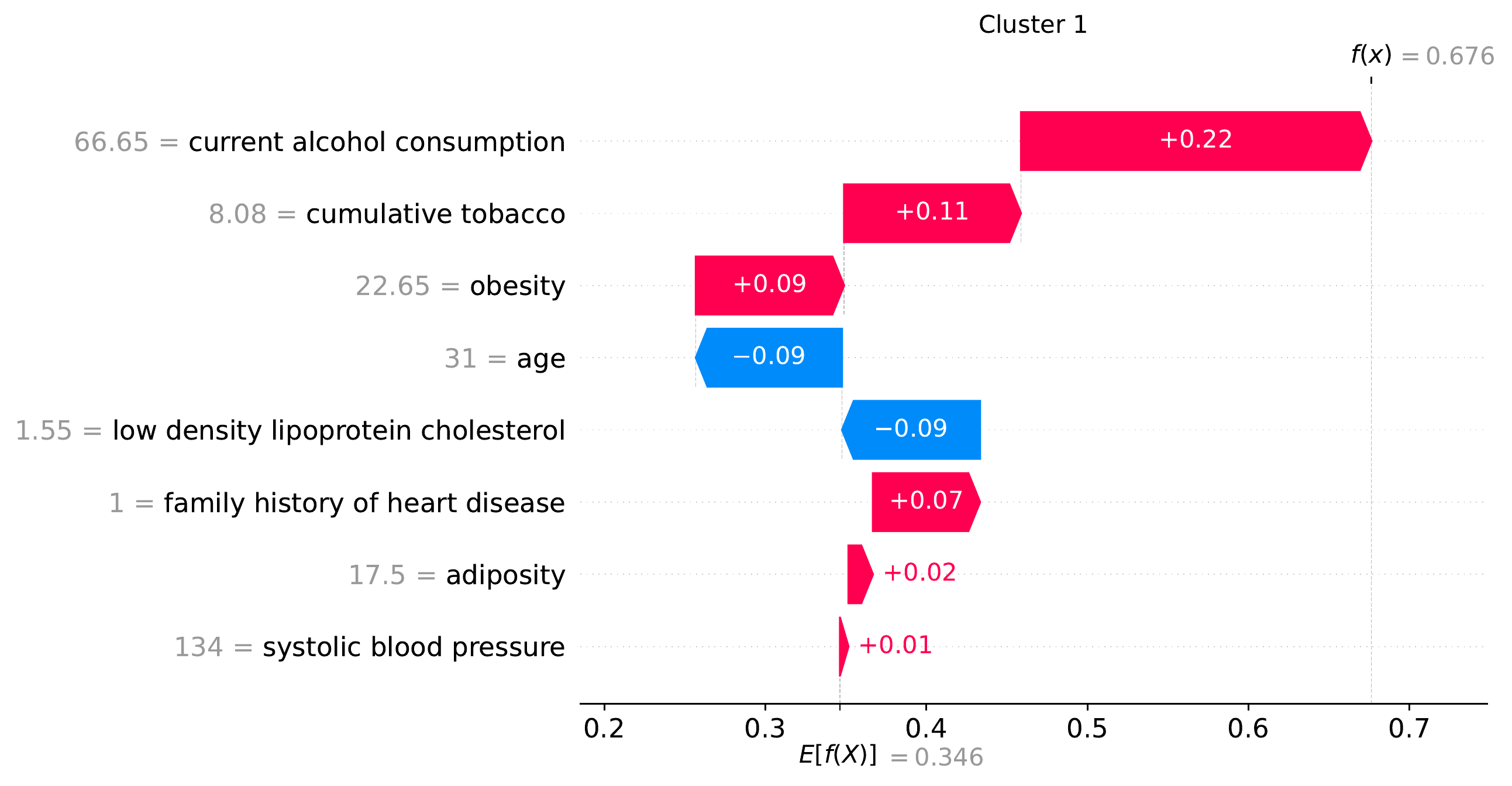}
\includegraphics [width=0.49\columnwidth]{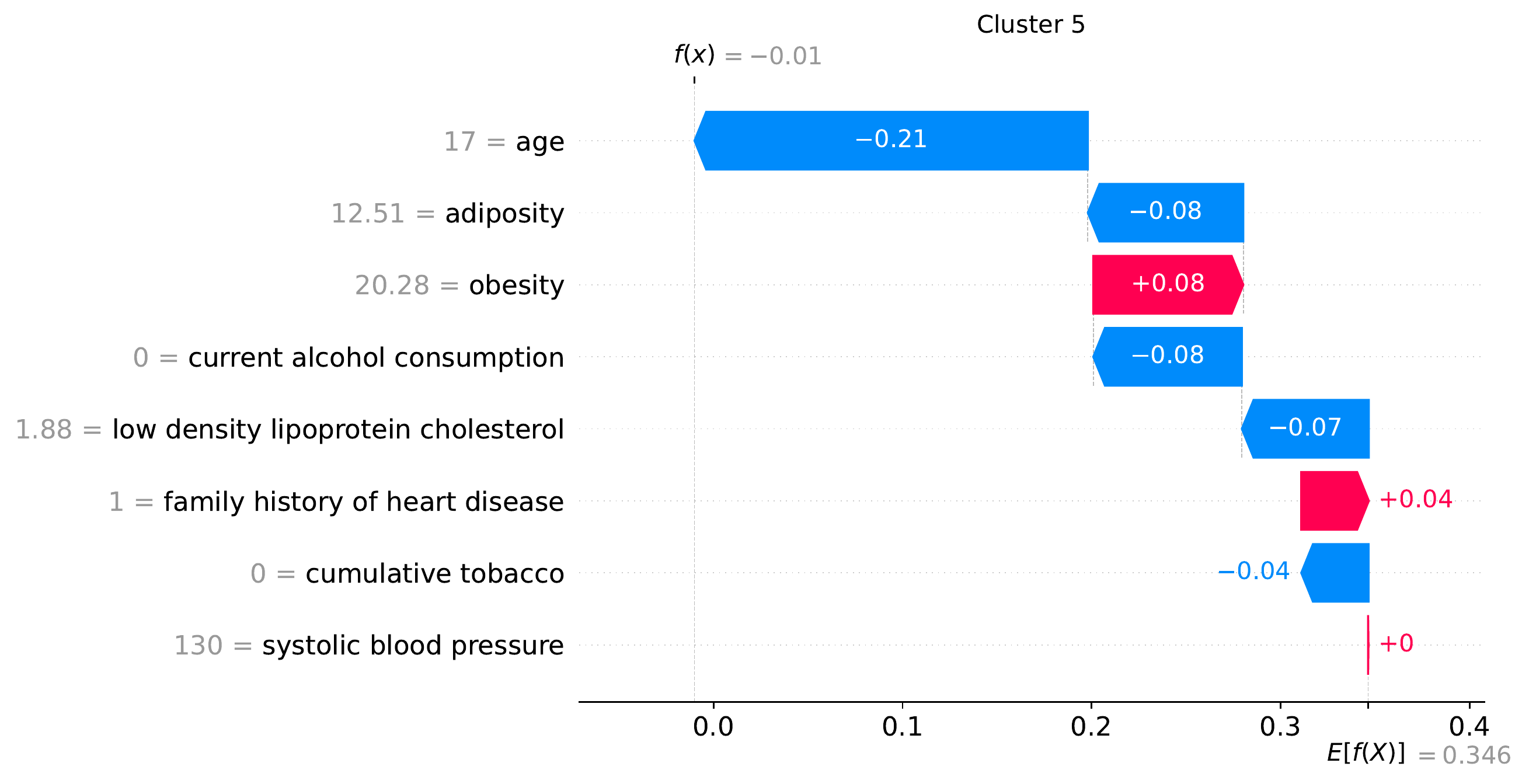}
\caption{Instances recommandées pour le cluster 1  à gauche et le cluster 5 à droite}
\label{fig:reco_inst}
\end{figure}

Avec ces clusters, il est possible d'aller plus loin et d'extraire des instances représentatives comme en Figure \ref{fig:reco_inst}.
Pour l'instance du cluster 1 à gauche, on peut constater que le patient "médian" est âgé de 31 ans ce qui semble le rendre moins à risque mais sa consommation d'alcool ainsi que sa consommation de tabac augmentent son risque de maladie cardio-vasculaire. Pour l'instance du cluster 5 à droite, le patient a 17 ans et dans son cas, l'absence de consommation d'alcool et de tabac réduit ses risques de maladie cardio-vasculaire. Les deux patients ont des antécédents familiaux, et sachant que le deuxième patient est encore jeune, un expert métier pourrait suivre son évolution et éventuellement faire de la prévention concernant les comportements qu'on soupçonne à risque.


\section{Conclusion}
Dans cet article, nous proposons un pipeline de traitement centré autour des explications et qui a pour objectif l'exploration de données. Ce pipeline s'appuie sur un modèle d'apprentissage capable de capturer les relations statistiques complexes et nous permet de les observer au travers de sous-groupes de données et des instances qui les représentent.
Nous appliquons ce pipeline à un jeu de données de santé.
Nous montrons alors qu'il permet d'extraire des connaissances importantes pour le data scientist comme pour l'expert métier. Les connaissances extraites permettent de détecter des biais dans les données ou des faiblesses sur le modèle. On observe par ailleurs des tendances démographiques complexes centrées autour de la problématique du jeu de données.
Des travaux à court termes seraient d'appliquer ce pipeline sur d'autres jeux de données pour généraliser l'approche proposée. 
Pour aller plus loin, il faudrait comparer les résultats obtenus avec différentes combinaisons de modèles, algorithmes d'explicabilité et sélections de variables. Cela permettrait d'observer si les connaissances extraites sont similaires, complémentaires  ou contradictoires. 
Il serait intéressant d'observer l'impact des hyperparamètres des modèles d'apprentissages et des algorithmes d'explicabilité sur les résultats.
Les conclusions tirées par ces travaux pourraient permettre de déterminer automatiquement les choix pour les différents éléments du pipeline en fonction du jeu de données.

\bibliographystyle{unsrtnat}
\bibliography{biblio_exemple}

\end{document}